\documentclass{article}

\usepackage{PRIMEarxiv}

\usepackage[utf8]{inputenc} 
\usepackage[T1]{fontenc}    
\usepackage{hyperref}       
\usepackage{url}            
\usepackage{booktabs}       
\usepackage{amsfonts}       
\usepackage{nicefrac}       
\usepackage{microtype}      
\usepackage{lipsum}
\usepackage{fancyhdr}       
\usepackage{graphicx}       

\graphicspath{{media/}}     

\title{Machine learning models and facial regions videos for estimating heart rate: a review on Patents, Datasets and Literature
}

\author{
Tiago Palma Pagano\textsuperscript{1}
Lucas Lemos Ortega\textsuperscript{1},
Victor Rocha Santos\textsuperscript{1} \\ 
\textbf{Yasmin da Silva Bonfim\textsuperscript{1},}
\textbf{José Vinícius Dantas Paranhos\textsuperscript{1},}
\textbf{Paulo Henrique Miranda Sá\textsuperscript{1}} \\ 
\textbf{Lian Filipe Santana Nascimento\textsuperscript{1}},
\textbf{Ingrid Winkler\textsuperscript{2}},
\textbf{Erick Giovani Sperandio Nascimento\textsuperscript{1*}} \\ \\
\textsuperscript{1}Software Departament, SENAI CIMATEC University Center, Salvador 41650010, Brazil\\
\textsuperscript{2}Department of Management and Industrial Technology, SENAI CIMATEC University Center, Salvador 41650010, Brazil\\
\textit{\textsuperscript{*}Corresponding author: erick.sperandio@fieb.org.br} \\
}

\begin{document}
\maketitle

\begin{abstract}

Estimating heart rate is important for monitoring users in various situations. Estimates based on facial videos are increasingly being researched because it makes it possible to monitor cardiac information in a non-invasive way and because the devices are simpler, requiring only cameras that capture the user's face. From these videos of the user's face, machine learning is able to estimate heart rate. This study investigates the benefits and challenges of using machine learning models to estimate heart rate from facial videos, through patents, datasets, and articles review. We searched Derwent Innovation, IEEE Xplore, Scopus, and Web of Science knowledge bases and identified 7 patent filings, 11 datasets, and 20 articles on heart rate, photoplethysmography, or electrocardiogram data. In terms of patents, we note the advantages of inventions related to heart rate estimation, as described by the authors. In terms of datasets, we discovered that most of them are for academic purposes and with different signs and annotations that allow coverage for subjects other than heartbeat estimation. In terms of articles, we discovered techniques, such as extracting regions of interest for heart rate reading and using Video Magnification for small motion extraction, and models such as EVM-CNN and VGG-16, that extract the observed individual's heart rate, the best regions of interest for signal extraction and ways to process them.
\end{abstract}


\section{Introduction} \label{Introduction}

The heart rate is a vital human body signal that provides the monitoring of a person's health. Heart rate estimation is particularly important for monitoring users in a range of frequent situations, such as driving a vehicle \cite{16}, engaging in physical activities \cite{schneider2018heart}, working in hazardous conditions \cite{sharma2018differences}, and during investigative police interviews \cite{bertilsson2020towards}. 
Heart rate or its variability may be used to track and detect stress-related aspects \cite{16}, tiredness \cite{15}, emotions \cite{18}, health \cite{young2018heart}, and social behavior \cite{colasante2017resting}.

Heart rate is the number of times the heart beats blood in one minute \cite{8}. In general, heart rate ranges from 41 to 240 Beats Per Minute (BPM) \cite{2, 7}. While for a resting adult, the heart rate can range from 60 to 100 BPM \cite{7, 8, 11}. The ElectroCardioGram (ECG) is a test that measures the resting heartbeat rhythm \cite{MARTISECG} and can be used to diagnose the patient's heart health. PhotoPlethysmoGraphy (PPG), on the other hand, is a technique for measuring blood volume variations \cite{PPG}, commonly obtained by a wrist or finger oximeter.

Visual estimating, which is often conducted with a camera and is less costly and less intrusive to the user, is an alternative to traditional measurement approaches \cite{huang2020visual}. As a result, research into heart rate estimating through video has increased, as has the usage of PPG in different heart rate assessment devices.

Machine Learning (ML) allows computers to learn real-world concepts and their relationships through the experiences contained in the data. From the data are extracted the various factors with the characteristics that influence the outcome. Variations that cannot always be easily observed in the data, but that influence the results \cite{goodfellow2016deep}. ML has been used for many different purposes, such as 3D object generation, drug creation, pandemic studies, image processing, face detection, image to text translation, texture transfer, traffic control, noise removal in images, among other \cite{aggarwal2021generative}.

Deep Learning (DL) has been used to estimate heart rate from facial recordings \cite{10}. DL techniques are multi-layered Artificial Neural Networks (ANN) with great degrees of flexibility, enabling efficient and effective classification of a wide variety of circumstances. Using a dataset, generalization of problems is achievable by modifying the internal parameters of the network to reflect the desired structure. This enables the discovery of optimum combinations of complex feature data. This particular ability enables the development of autonomous systems capable of making human-like decisions.

There has been no prior study that has reviewed the benefits and constraints of using machine learning models to estimate heart rate from facial videos, , to the best of the authors' knowledge. Nonetheless, this comprehension might play a vital role in the development of applications in this sector. Furthermore, patents give strategic information to industry and academics, as well as a resource for technical management and innovation. As a consequence, patent examination may specify a particular technology, allowing for the discovery of technology advancements, inventors, market trends, and other features. 

Several Virtual Reality and Augmented Reality devices are already equipped with cameras that capture the user's face; these can be leveraged for heart rate estimation and monitoring the user's condition in certain contexts \cite{11}.

Thus, this study aims to analyze the benefits and challenges of using machine learning models to estimate heart rate from facial videos, through patents, datasets and articles reviews.

This paper is organized into four parts. Section \ref{Materials and Methods} outlines the methods used, Section \ref{Results and discussion} details our findings, and Section \ref{Conclusions} presents our final considerations and recommendations for future investigations.

\section{Materials and Methods} \label{Materials and Methods}

This patent, dataset, and literature systematic review adopted the Preferred Reporting Items for Systematic Reviews and Meta-Analyses (PRISMA) standards, which were developed to “help systematic reviewers transparently report why the review was done, what the authors did, and what they found” \cite{page2021prisma}. Besides, the method described in \cite{booth2016systematic} was used, which comprises seven stages: planning, defining the scope, searching the literature, assessing the evidence base, synthesizing, analyzing, and writing. The preliminary search strategy was designed by a team of five machine learning model developers and was then assessed by three senior ML researchers. The developed method is explained in the following sections.

\subsection{Planning}

The knowledge bases that will be examined are determined during the planning stage \cite{booth2016systematic}. We chose the Derwent Innovation Index database for patent search because of its vast coverage: it comprises 39.4 million patent families and 81.1 million patent data, as well as 59 foreign patenting bodies and two journal sources. The Derwent platform includes distinct features that increase data extraction, such as the "Smart Search" feature, which uses artificial intelligence to boost keyword finding. \cite{grames2019automated}.

Another tool is the Derwent World Patents Index (DWPI), the world's most extensive database of enhanced patent information, which includes extended patent titles and abstracts, English abstracts of the original patent, and an advanced categorization system \cite{codes2018derwent}.

We used Google's indexer to perform an exploratory search of face video datasets to be used for heart rate estimation since there are no dedicated databases for searching datasets.

Scopus, Web of Science, and IEEE Xplore were used to search for articles. They were chosen because they are worldwide multidisciplinary scientific databases with extensive citation indexing coverage. Scopus now includes 81 million inspected documents, Web of Science has over 82 million items, and IEEE Xplore has about 5 million documents.


\subsection{Defining the Scope}

The defining the scope stage is focused with well stated research questions \cite{booth2016systematic}. We determined three key research questions to investigate, which are as follows:

Q1: How are patents on using machine learning models to estimate heart rate from facial videos characterized, in terms of assignees, publications per year, and advantages of the invention?

Q2: How are available datasets in ANN training for heart rate estimation characterized?

Q3: How are current knowledge on the use of machine learning models to estimate heart rate from facial videos characterized, in terms of use cases, methodologies and models, and metrics comparison? 


\subsection{Searching the Literature} \label{Searching the Literature}

This stage comprises scanning the databases indicated in the planning stage for a phrase relevant to the research questions specified in the defining the scope stage \cite{booth2016systematic}.

We searched through the databases using the following potential search phrase:

\begin{center}
((( ``heart rate'' OR ``heart activity'' OR ``heart rate estimation'' ) AND ( ``vital signs'' OR ``remote ppg'' OR ``remote photoplethysmography'' ) AND ( ``Image'' OR ``Video'' ) AND ( ``face'' OR ``facial'')))
\end{center}

This potential search phrase was then fed into the Biblioshiny tool for bibliometrics analysis. As shown in Figure \ref{FIG:rede_de_co_ocorrencia}, the co-occurrence network of terms revealed ``PPG'', ``heart'' and ``heart rate'' as the most key elements of the two clusters created.

\subsection{Assessing the Evidence Base}

The preliminary search is adjusted at this stage, and inclusion and exclusion criteria filters are used to select the relevant records found during the searching the literature stage  \cite{booth2021systematic}, so the LitSearch library was adopted to optimize the proposed initial search phrase. With this, the 99 articles were used from the initial search and the most relevant keywords were defined by the LitSearch tool and the search phrase below was generated.

\begin{center}
( ``blood volume''  OR  ``health monitoring''  OR  ``signal processing''  OR  ``vital sign'' )  AND  ( ``facial region''  OR  ``facial expression''  OR  ``imaging photoplethysmography''  OR  ``facial video'' )  AND  ( ``heart''  OR  ``volume pulse''  OR  ``heart rate'' )  AND  ( ``convolutional neural network''  OR  ``deep learning''  OR  ``neural network'' )
\end{center}

We limited the amount of records returned at the previous stage by applying the following exclusion criteria:

E1: Exclude patents filed or articles published before to 2016;
E2: Exclude items that are not written in English.
E3: Exclude dead patent applications.

Similar search keywords were used to obtain articles and datasets, with minor changes to meet the search engine parameters of Scopus, Web of Science and IEEE Xplore databases.

The search was conducted in May 2021, and we screened 7 patent records, 11 datasets, and 20 articles after applying exclusion criteria.

\subsection{Synthetising and Analysing}
\label{section:sintese}

Derwent analytical and "Insights" tools were used to scrutinize the patents. The recovered items were exported to the Mendeley Reference Manager tool, and Microsoft Excel visuals were developed.

\begin{figure*}[ht]
	\centering
		\includegraphics[scale=.8]{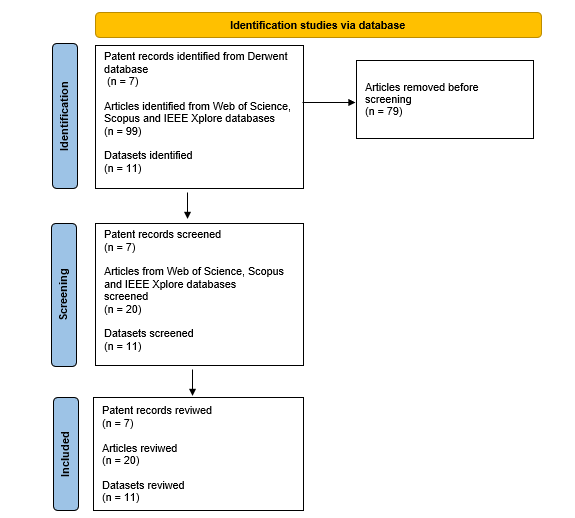}
	\caption{Systematic review flow diagram, adapted from PRISMA 2020.}
	\label{FIG:prisma_diagram}
\end{figure*}

Figure \ref{FIG:prisma_diagram} represents the stages of this systematic review.

\section{Results and discussion} \label{Results and discussion}

The research questions Q1, Q2, and Q3 were examined in order to discover the benefits and limitations of employing ML for heart rate estimation from facial video. In the sections that follow, we analyze our findings.

\subsection{Patents landscape}
Table \ref{Table:tabelapat} shows the seven patent records identified.

To address the first research question, these patents were analyzed to answer frequent concerns and uncover patterns in assignees, publications per year, and benefits.

Q1: How are patents on using machine learning models to estimate heart rate from facial videos characterized, in terms of assignees, publications per year, and advantages of the invention?

The characterization of patent assignees may contribute in the identification of industry leaders, the assessment of potential competitors, and the identification of niche players. Except for GB2572961A, we discovered that the majority of published patents are from Chinese applicants. Furthermore, the widely dispersed distribution of publishe patent among multiple assignees is interesting: just one assignee, Hangzhou First People's Hospital, submitted two patents, whilst the other patents were filed by distinct assignees.

Instead of a relatively equal-sized but vast portfolio held by a few organizations, indicating an active competitive market with strong investments by multiple companies, implying that the market is difficult to enter, we discovered a large number of assignees, each with a small number of records, indicating a developing technology space. Acquisition or quick development may be used to enter this area. There are multiple companies, each with a low number of patents, signaling a chance to approach this field while it is still in its inception, either by licensing existing technology, acquiring one of the competitors, or inventing new technology that is not currently patented. It is also important to note that three of the seven assignees are from the health sector, and three of the seven assignees are universities.

We found a substantial increase in annual patent publication in 2020. There were no patents published until 2018, after which one was published in 2019, and six were published in 2020. Because of the 18-month patent legal confidentially rule, we excluded 2021 patents from our analysis. 

In terms of benefits, we examined the seven patents in relation to the advantages of the invention as described by the authors and the novelty of the invention, i.e. the unique innovative feature introduced by the inventor that is not conventional and an improvement on existing technology.

\begin{table*}[ht]
\centering
\resizebox{\textwidth}{!}{%
\begin{tabular}{cp{8cm}p{3cm}}
\hline
\centering
    Identification Number  &
    \centering
    Title &
    Assignee \\ \hline
    WO2019202305A1  &
    \centering
    System for vital sign detection from a video stream &
    ClinicCo Ltd \\ \hline
    CN110738155A &
    \centering
    Face recognition method and device, computer equipment and storage medium  &
    Hangzhou First People's Hospital \\ \hline
    CN110909717A &
    \centering
    Moving object vital sign detection method &
    Nanjing University of Science and Technology \\ \hline
    CN111259787A &
    \centering
    Unlocking method and device, computer equipment and storage medium &
    Hangzhou First People's Hospital \\ \hline
    CN111260634A &
    \centering
    Facial blood flow distribution extraction method and system &
    Tianjin Polytechnic University \\ \hline
    CN111797794A &
    \centering
    Facial dynamic blood flow distribution detection method &
    People's Public Security University of China \\ \hline
    US20200155040A1 &
    \centering
    Systems and methods for determining subject positioning and vital signs &
    Hill Rom Services Inc \\ \hline

\end{tabular}%
}
\caption{List of retrieved patents.}
\label{Table:tabelapat}
\end{table*}

The patent US2020155040-A1 proposes a method for automatically monitoring a patient's position and vital signs. Using Near-InfraRed (NIR) cameras and Long Wavelength Infrared (LWIR) to monitor hospital patients by measuring heart rate and facial temperature. 

The patent CN111797794-A proposes a method to detect the flow of blood distribution on people's faces Video of the face is captured with a Red Blue Green (RGB) camera to determine a Region of Interest (ROI), ROI heart rate pattern via Remote PhotoPlethysmoGraphy (rPPG) and obtain blood flow from each sub-region. The value of blood flow intensity is obtained from the pulse wave of each sub-region. In addition, the distribution of facial blood flow can be obtained from the value of the blood flow intensity identified in each sub-region. This process does not require subject skin contact to be tested and is highly accurate. 

The patent CN110909717-A aims to detect the vital signs of a moving person by performing face recognition, as well as obtaining the heart rate by displaying the information in real-time. The vital sign is extracted from the subject’s face and it is detected with help of face detection, however, if it could not be detected the detection of vital signs can be failed. In addition, the heart rate values can be obtained by vital signs, and caching and displaying the heart rate value in real-time. This method reduces the complexity of the measurement and improves the accuracy of the signal extraction.

A method for extracting the facial blood flow distribution is proposed by the patent CN111260634-A. The video of the person's face is captured with an RGB camera, where each frame of the video is divided into sub-regions, in this way, the pulse rate is determined in each separate sub-region using an rPPG algorithm. The pulse wave signal of the sub-regions of a frame is obtained by rPPG and its signal is used as the value of the blood flow distribution, allowing identified facial blood flow distribution. In addition, the heart rate frequency is determined by the maximum amplitude in a spectrogram. 

The patent CN111259787-A proposes a method to unlock doors or devices with heart rate and state of inebriation based on data from a person's face captured by a camera, aiming to enhance biometric security. Face recognition, heart rate detection and drunk detection are performed based on face information. In addition, human face image and human face temperature information are also obtained to aid in heart rate and drunk detection. Face recognition and heart rate detection are performed based on the face image, which allows you to get the results of face recognition and heart rate detection.

 The patent CN110738155-A proposes a method of facial recognition that includes capturing the head and the entire body, applying image amplification with Eulerian Video Magnification (EVM) to extract the original signal and heart rate. Human face information and human heart rate information are obtained based on face image and off face image. This method avoids the influence of external light and body detection, resulting in an accurate answer for face recognition. In addition, the human heart rate information is obtained based on the rate information environment. 
 
 Finally, the patent WO2019202305-A1, a method of detecting a user's heart rate from transmitted video is proposed. The face is detected in the frames and your motion is analyzed on the face detection. The heart rate is determined by changing the colours of pixels and face movement between the video motion. Additionally, one heart rate of the subject is determined based on a selected one of the first and second estimated pulse frequencies.

\subsection{Datasets} 

Table \ref{Table:tabeladatasets} shows the eleven datasets selected by the search strategy. 

These eleven datasets were analyzed to address the second research question:

Q2: How are available datasets in ANN training for heart rate estimation characterized?

We found out that majority of datasets serve academic objectives and come with a variety of licenses.

The MMSE-HR (Multimodal Spontaneous Expression - Heart Rate) dataset \cite{MMSE} is composed of their facial emotions annotated in terms of occurrence and intensity. In addition, each video's associated heart rate and blood pressure sequences are included in the dataset. It comes in three different licensing types: standard, individual, and corporate. Because of the large number of participants, the dataset may have a benefit over the others in terms of heart rate estimate.

VIPL-HR \cite{VIPL} includes videos of people in nine different situations of head movement and lighting, as well as their heart rates.
The dataset is exclusively accessible for academic study and needs the completion of a release form. As recorded in Table \ref{Table:tabeladatasets}, the dataset has the advantage of incorporating infrared captured facial films and the largest number of individuals among the examined datasets.

The DEAP \cite{DEAP} is an emotion analysis dataset that employs electroencephalogram, physiological, and visual signals. It is broken into two sections. In the first, the participants rate segments of music videos on arousal, valence, and dominance. In the second experiment, the participants watch and rate the same movies as before while their electroencephalogram and physiological data are recorded. In addition, RGB videos of participants' faces were captured. The dataset is only permitted for use in academic research, and form submission is needed for the study's purposes.

The COHFACE dataset \cite{COHFACE} contains RGB videos of people's faces in 480p and 20 frames per second, as well as their blood volume and breathing rate pulses. It is necessary to fill out a form with justification in order to obtain access to it. The EULA license covers the dataset.

The MAHNOB \cite{MAHNOB} features RGB videos of people's faces in six different poses, as well as their ECG, ElectroEncephaloGram (EEG), breath amplitude, and skin temperature data. The dataset is protected under the EULA license and is available for download on the website.

The ECG-Fitness dataset \cite{ECG} includes 1080p RGB films of persons undertaking physical exercises on fitness equipment, as well as their ECG signals. It is free for academic usage and requires the signing of a compliance agreement. By presenting six situations among its videos, the dataset includes a greater variety of heart rates.

The R-PPG Algorithm performance dataset \cite{PBDT} is used to assess the robustness of rPPG against changes and high fluctuations in heart rate and pulse rate. It includes RGB videos of subjects as well as their distinct ECG signals. The dataset is free to download and is licensed under the 4TU general terms of use. In comparison to previous datasets, R-PPG Algorithm has fewer participants and no infrared video, which can be observed in Table \ref{Table:tabeladatasets}.

The MERL-Rice Near-INfrared Pulse (MR-NIRP) \cite{NIRP} is a publicly available dataset that is separated into two contexts: within an automobile and inside a room. The MR-NIRP Vehicle has films of the faces of individuals in a car, for a total of ten experiments per person. The MR-NIRP Indoor is recorded in a room, with each of the eight participants' faces recorded either static or moving. The contexts include files containing data from a pulse oximeter connected to the participants' fingers. Sixteen of the participants are males, four of them have beards, and two are women. They are all between the ages of 20 and 60, with Asian, Indian, and Caucasian complexion tones. The only requirement for utilizing datasets is to reference them in publications.

The Imaging Photoplethysmography Dataset \cite{HRV} is a public dataset that contains RGB videos (1920x1200) of the faces of participants as well as their PPGs and BPMs. The dataset is freely accessible.

The Toadstool \cite{ToadstoolUrl} dataset is intended for academic usage only, although it may be used commercially with a license subscription. It is licensed under the Creative Commons Attribution-NonCommercial 4.0 International (CC BY-NC 4.0) license and includes 480p videos of the faces of participants playing Super Mario Bros, as well as heart rate data, blood volume pulse, and beat intervals extracted from an Empatica E4 wristband.

The dataset Univ. Bourgogne Franche-Comté Remote PhotoPlethysmoGraphy (UBFC-RPPG) \cite{UBFC} comprises 480p videos of participants' faces and their individual pulse signals.

\begin{table*}
\centering
\begin{tabular}{cccc}
\hline
\centering
Dataset  &   Number of videos &  Subjects &  Has infrared videos \\ \hline
    MMSE-HR &
    \centering
    102 &
    40 & 
    No \\ \hline
     VIPL-HR &
    \centering
    3230 &
    107 & 
    Yes \\ \hline
    Deap - Part 01 &
    \centering
    120 &
    14-16 & 
    No \\ \hline
    Deap - Part 02 &
    \centering
    40 &
    32 &
    No \\ \hline
    COHFACE &
    \centering
    160 &
    40 &
    No \\ \hline
    MAHNOB &
    \centering
    3741 &
    27 & 
    Yes\\ \hline
    ECG-Fitness dataset &
    \centering
    207 &
    17 &
    No\\ \hline
    R-PPG &
    \centering
    21 &
    03 &
    No\\ \hline
    MR-NIRP &
    \centering
    180 &
    18 & 
    Yes \\ \hline
    The Imaging Photoplethysmography Dataset &
    \centering
    60 &
    12 &
    No\\ \hline
    Toadstool &
    \centering
    10 &
    10 &
    No \\ \hline
    UBFC-RPPG &
    \centering
    42 &
    42 &
    No \\ \hline
\end{tabular}
\caption{List of retrieved datasets.}
\label{Table:tabeladatasets}
\end{table*}

As previously stated, there is a vast range of datasets accessible, each with a unique participant distribution and information architecture.



\subsection{Articles landscape}

Table \ref{Table:tabelaart} lists the twenty articles that were selected. These articles were reviewed in order to answer the research question Q2, and our findings are provided in the subsections that follow:

Q3: How are current knowledge on the use of machine learning models to estimate heart rate from facial videos characterized, in terms of use cases, methodologies and models, and metrics comparison?

\begin{table}[ht]
\centering

\begin{tabular}{p{.3cm}p{12cm}p{2.3cm}}
\hline
Item &  Title & Publication Year\\ \hline
1 & Emotion recognition from facial expressions and contactless heart rate using knowledge graph \cite{18}. & 2020 \\ \hline
2 & Toadstool: A dataset for training emotional intelligent machines playing Super Mario Bros \cite{Toadstool}. & 2020 \\ \hline
3 & A deep learning framework for heart rate estimation from facial videos \cite{2}. & 2020 \\ \hline
4 & Non-contact-based driver's cognitive load classification using physiological and vehicular parameters \cite{16}. & 2020 \\ \hline
5 & Non-Contact Emotion Recognition Combining Heart Rate and Facial Expression for Interactive Gaming Environments \cite{8}. & 2020 \\ \hline
6 & Visual Heart Rate Estimation from Facial Video Based on CNN \cite{3}. & 2020 \\ \hline
7 & On assessing driver awareness of situational criticalities: Multi-modal bio-sensing and vision-based analysis, evaluations, and insights \cite{15}. & 2020 \\ \hline
8 & DeepPerfusion: Camera-based Blood Volume Pulse Extraction Using a 3D Convolutional Neural Network \cite{10}. & 2020 \\ \hline
9 & Heart Rate Estimation from Facial Videos Using a Spatiotemporal Representation with Convolutional Neural Networks \cite{11}. & 2020 \\ \hline
10 & Robust remote heart rate estimation from face utilizing spatial-temporal attention \cite{12}. & 2019 \\ \hline
11 & Automatic Monitoring of Driver's Physiological Parameters Based on Microarray Camera \cite{17}. & 2019 \\ \hline
12 & Combating the impact of video compression on non-contact vital sign measurement using supervised learning \cite{5}. & 2019 \\ \hline
13 & Architectural tricks for deep learning in remote photoplethysmography \cite{14}. & 2019 \\ \hline
14 & Emotion inference of game users with heart rate wristbands and artificial neural networks \cite{19}. & 2019 \\ \hline
15 & EVM-CNN: Real-Time Contactless Heart Rate Estimation from Facial Video \cite{4}. & 2019 \\ \hline
16 & Long Distance Vital Signs Monitoring with Person Identification for Smart Home Solutions \cite{7}. & 2018 \\ \hline
17 & A Novel Short-Term Event Extraction Algorithm for Biomedical Signals \cite{9}. & 2018 \\ \hline
18 & Deep learning with time-frequency representation for pulse estimation from facial videos \cite{1}. & 2018 \\ \hline
19 & Deep super resolution for recovering physiological information from videos \cite{6}. & 2018 \\ \hline
20 & Towards Generic Modelling of Viewer Interest Using Facial Expression and Heart Rate Features \cite{13}. & 2018 \\ \hline
\end{tabular}

\caption{List of retrieved articles.}
\label{Table:tabelaart}
\end{table}

\subsubsection{Bibliometric analysis} 

\begin{figure*}[ht]
	\centering
		\includegraphics[scale=.65]{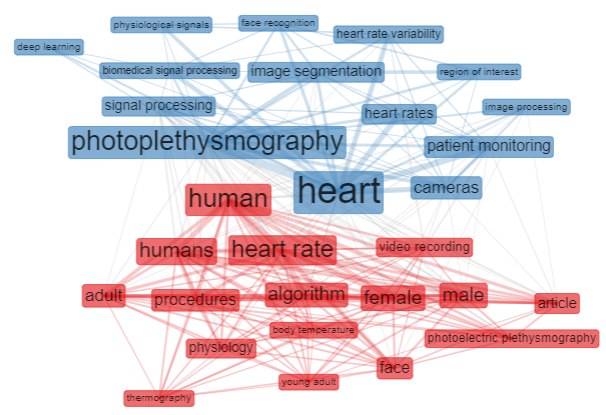}
	\caption{Keyword co-occurrence network}
	\label{FIG:rede_de_co_ocorrencia}
\end{figure*}

Figure \ref{FIG:rede_de_co_ocorrencia} depicts the total extent of the key terms found in the reviewed articles. The terms are clearly divided into two groups: the blue group refers to terms that we will use to define the technical scope of the bibliometric analysis, while the red group identifies the research objectives associated with the technical terms.

\begin{figure*}[ht]
	\centering
		\includegraphics[scale=.30]{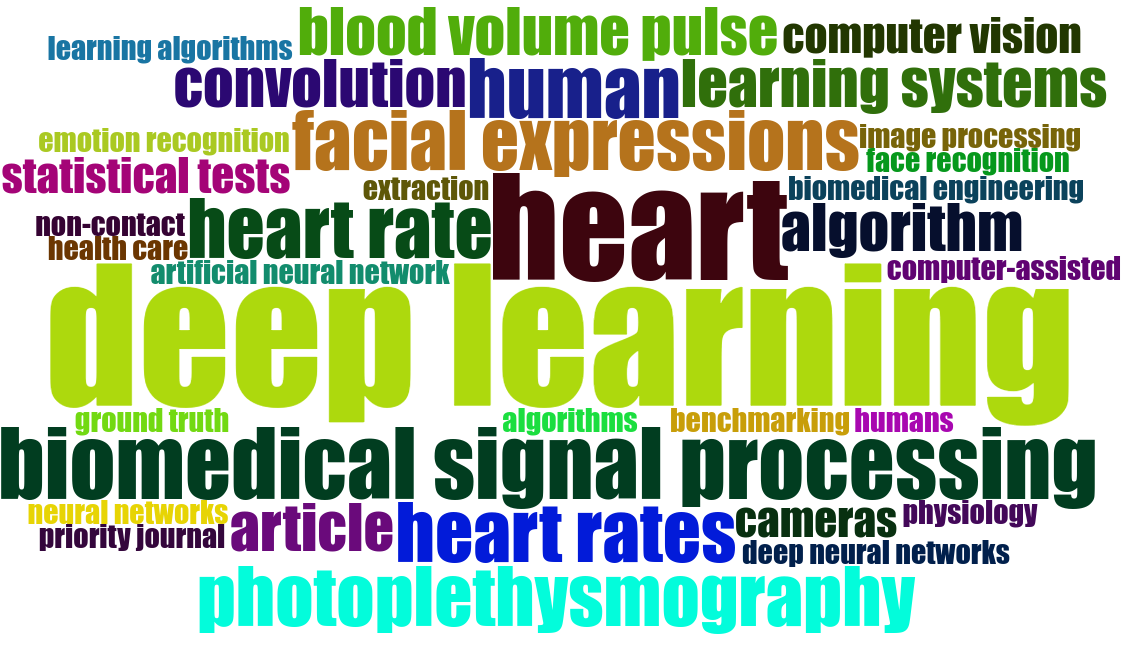}
	\caption{Search Keyword Cloud}
	\label{FIG:Nuvem_de_palavras}
\end{figure*}

Figure \ref{FIG:Nuvem_de_palavras} depicts the extracted word cloud from the reviewed articles, indicating which terms are most related to the subject. The terms with the greatest incidence numbers are connected to cardiac monitoring using cameras that display a human face image. Other terminology used to describe comparable approaches include image segmentation, face recognition, image and signal processing, independent component analysis, and others.

\begin{figure*}[ht]
	\centering
		\includegraphics[scale=.55]{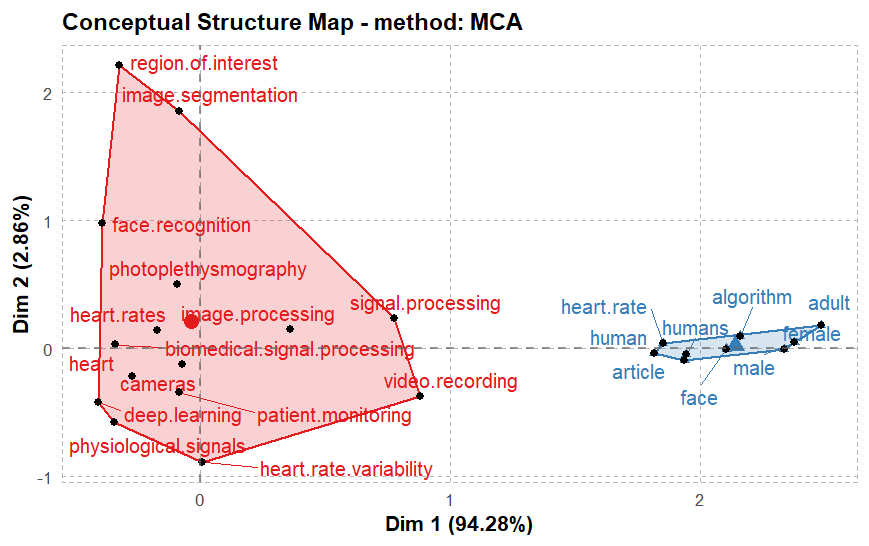}
	\caption{Conceptual Framework Map}
	\label{FIG:estrutura_conceitual}
\end{figure*}

The conceptual structure of the reviewed articles is shown in Figure \ref{FIG:estrutura_conceitual}. This structure covers almost 97 percent of the primary subjects mentioned in the articles. The clusters represent how ideas are connected; the closer they are, the stronger the association. There is a cluster of topics about techniques and another about applications. Figure \ref{FIG:estrutura_conceitual} demonstrates the density that the technical terms in Figure \ref{FIG:rede_de_co_ocorrencia} have in relation to the terms that refer to the object of study.

\begin{figure*}[ht]
	\centering
		\includegraphics[scale=.55]{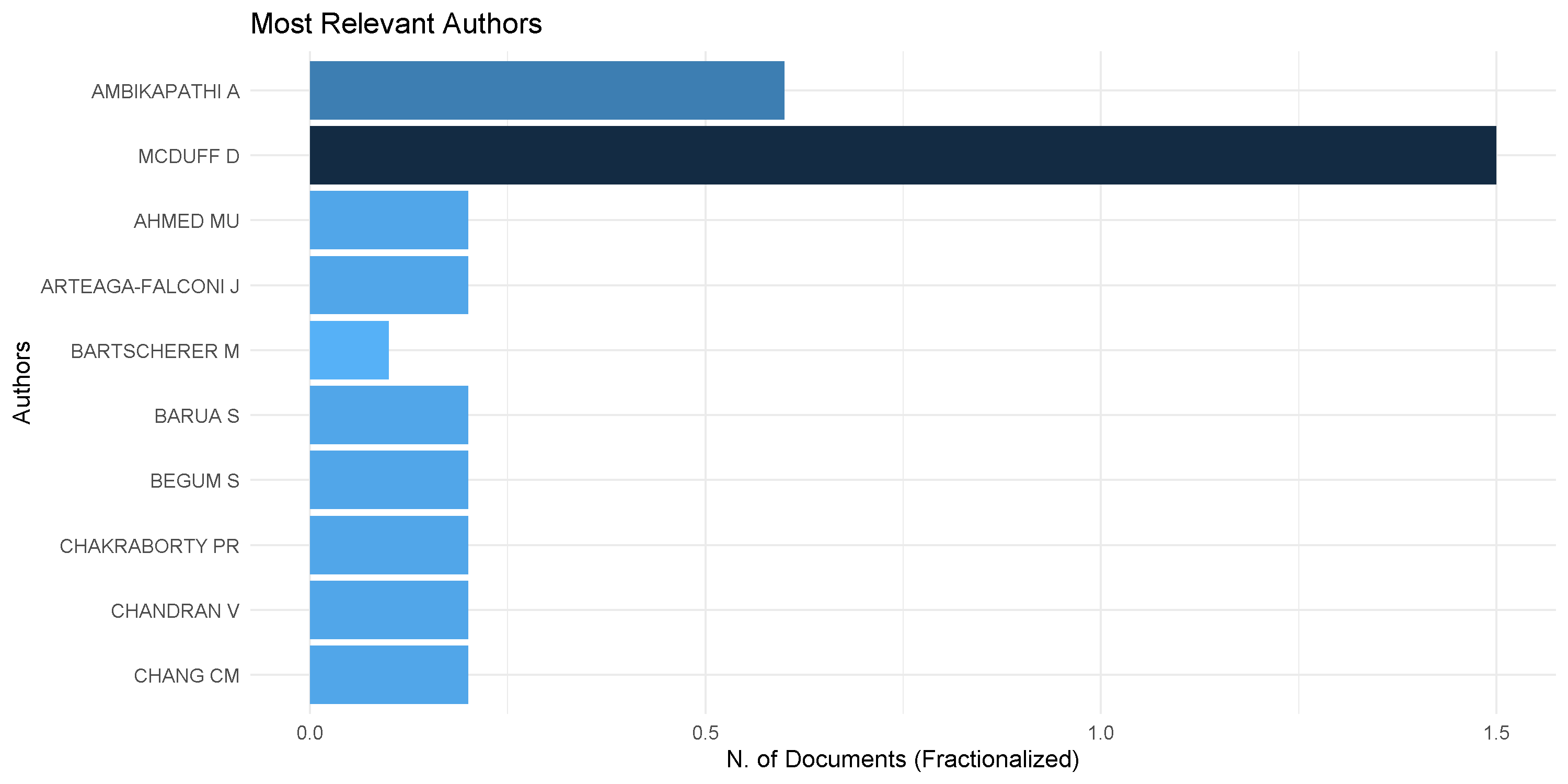}
	\caption{The impact of authors}
	\label{FIG:Impacto_dos_autores}
\end{figure*}

\begin{figure*}[ht]
	\centering
		\includegraphics[scale=.65]{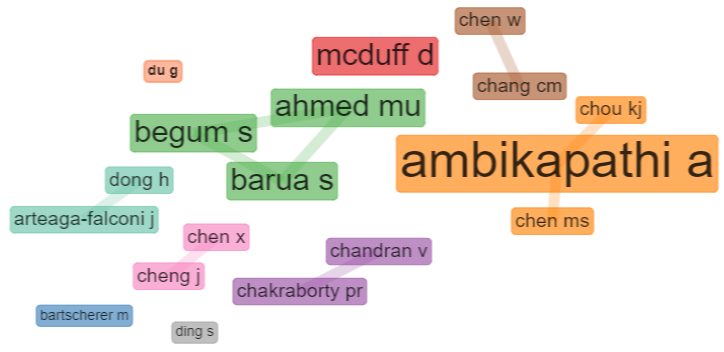}
	\caption{Collaboration Network}
	\label{FIG:rede_de_colaboracoes}
\end{figure*}

The Figure \ref{FIG:Impacto_dos_autores} illustrates the writers who had the biggest impact on the research theme. Among them, the authors Ambikapathi, A., and McDuff, D. stand out as having the greatest relevance among those selected for this review. It can be seen that although McDuff, D. shows the highest relevance in Figure \ref{FIG:Impacto_dos_autores}, Ambikapathi, A. shows the largest name in Figure \ref{FIG:rede_de_colaboracoes}, and therefore, higher impact in the thematic group.


The Figure \ref{FIG:rede_de_colaboracoes} shows the collaborative networks of authors found in the publications, in addition, the longer the author's name, the greater its appearance in the works found. The division by color reveals the group of authors who worked on the same work and the size of the node indicates the impact on the thematic group of the research.

\subsubsection{Use cases}

\cite{1, 2, 4, 7, 9, 11, 14, 17, 18} reduced the ROIs of the face using facial landmark detectors, while \cite{3, 5, 6, 8, 10, 12} utilized the full face as model input. In \cite{1, 2}, they performed removal of irrelevant signals such as tremor, illumination changes, and so on in order to locate and cut out ROIs to estimate heart rate. \cite{4} demonstrated that utilizing the whole face or ROIs offers positive results for heart rate estimation, but both have advantages and disadvantages. When the whole face is used, there is extra processing to clean up the signals. When there is ROI clipping, \cite{4} pointed out that, despite requiring less processing in terms of treating irrelevant regions, the challenge of providing a consistent detection of face landmarks to smooth noisy signals occurred. \cite{1, 2, 3, 4, 5, 6, 7, 8, 9, 10, 11, 12, 14, 17, 18} estimated heart rate using face videos, while \cite{8, 13, 15, 16, 19} estimated additional physiological signals using contact devices or heart rate estimation techniques. 

Using case studies, several previously investigated methodologies, as well as their benefits and disadvantages, are identified, enabling the search for alternative answers to challenges to be deeper without the need to repeat prior mistakes or commit time to research a path already taken.

\subsubsection{Models}

There is a wide variety of adopted techniques used for data capture, data processing, and final estimation. Most of the articles used Convolutional Neural Networks (CNNs) to build ANNs for noise treatment in order to perform heart rate estimations.

\cite{3, 8} collected temporal heart rate information using recurrent layers for Long and Short Term Memory (LSTM).
Furthermore, \cite{7, 9, 17} employed other techniques to obtain heart rate. For face detection of ROIs, \cite{10, 14} employed Opencv, while \cite{1, 2} used a Single Shot Multibox Detector (SSD) network. ROIs are utilized to monitor color changes at places where heart rate estimate is achieved by viewing such changes. \cite{7} employed EVM as an amplifier of the raw signal, which was then filtered with the Butterworth bandpass filter (0.67 Hz - 4 Hz), a filter used by \cite{9, 10} that positions the heart rate found in the optimal ranges for the analysis of the issue. The heart rate was extracted from films acquired by RGBs cameras, the most popular and widely available on the market, in \cite{1, 6, 8, 9, 11}. \cite{6, 17, 18} employed Independent Component Analysis (ICA) to allow independent signal analysis, and \cite{9}, used the Relative Energy (Rel-En) algorithm, which determines the heart rate given the preprocessed signal from the ROI's green channel.

In \cite{17}, the algorithm Joint Approximation Diagonalization Estimation of Real Signals (JADER) and ICA were used to separate the RGB components and determine the heart rate. As a result, most studies conducted preprocessing of the extracted signals, mostly for heart rate normalization, while \cite{3, 5} did not undertake preprocessing of the face videos.

Among the multiple methodologies used to validate ANN estimations, \cite{7, 11, 12, 18} evaluated heart rate estimation outcomes using mean error calculation metrics such as Root Mean Square Error (RMSE), Mean Absolute Error (MAE), Mean Error Rate (MER), and L1 loss. Furthermore, \cite{11, 12} used Pearson's correlation coefficient to support the assessment of the estimated results by comparing the estimated data to the ground truth data. \cite{11, 12, 18} also examined the Standard Deviation (SD\textsubscript{e}) to analyze the dispersion of the estimations.

The preferred use of the green channel in \cite{1, 2, 17} studies and the filtering of external influences to the recordings are two key and recurring themes in several articles.
The former is because it is more dependable when it comes to extracting physiological signs from the individual in the video. The latter, on the other hand, is more detrimental to estimation since it employs techniques such as Joint Blind Source Separation (JBSS), feature-decoders, and Gaussian filters. The strategy of approaching heartbeat estimation as a classification problem in \cite{14} resulted in improved network performance in estimating heart rate from videos with dynamic circumstances. In contrast, it was discovered in \cite{15} that Heart Rate Variability (HRV) is an excellent metric for classifying cognitive states and is also more robust than heart rate.

\subsubsection{Comparison of metrics}

\cite{1} used the MAHNOB-HCI dataset during the test. According to the authors, the study surpasses other works in the RMSE, SD\textsubscript{e}, and MER metrics, but the Mean Error (M\textsubscript{e}) and Pearson Correlation metrics yielded similar results.

\cite{2} used the dataset MAHNOB-HCI during the test and stated that the technique outperformed \cite{1} in the RMSE, Pearson Correlation, and SD\textsubscript{e} metrics, and was in the top three in the others. During the test with the dataset VIPL-HR, and according to \cite{2}, the technique surpassed previous studies in terms of RMSE while placed in the top three in the other metrics.

\cite{3} employed their dataset. The strategy outperformed the competition in the Pearson Product Moment Correlation (PPMC) metric and was in the top three in the SD\textsubscript{e} and RMSE metrics. With the exception of PPMC, EVM-CNN \cite{4} outperformed all other approaches in this study.

\begin{figure*}[ht]
	\centering
		\includegraphics[scale=.65]{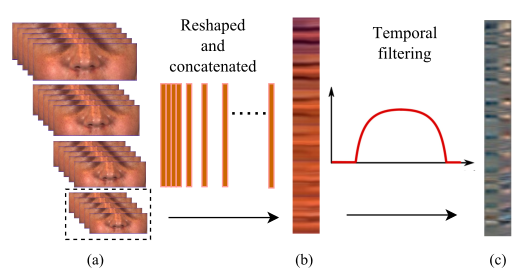}
	\caption{EVM-CNN \cite{4} process of extracting the feature images.}
	\label{FIG:evm_feature_image}
\end{figure*}

\cite{4} outperformed the others on all criteria using the MMSE-HR dataset, the process of building the feature images is illustrated in Figure \ref{FIG:evm_feature_image}. Using the MAHNOB-HCI dataset, the study obtained the same performance metrics as the others, outperforming them on all criteria.

When the findings from the MAHNOB-HCI dataset of \cite{4} were compared to those of \cite{2} and \cite{1}, \cite{4} exceeded all measures except RMSE by 0.18 points.

During testing with the \cite{estepp2014recovering} dataset, \cite{5} produced the following findings: (1) the greater the video's Compression Rate Factor (CRF), the higher the MAE and the lower the Signal-to-Noise Ratio (SNR), i.e. the worse the quality of heart rate estimate; (2) networks trained and tested with the same CRF had better metrics as compared to results from training with a CRF lower than the test CRF.

During testing with the dataset of \cite{estepp2014recovering}, \cite{6} proved that the super-resolution preprocessing network with the Image PhotoPlethysmoGraphy (iPPG) ICA, the network outperformed standard upsampling methods on all measures except SNR. This illustrates how utilizing a super-resolution network as a preprocessing step for low-resolution videos may improve the accuracy of heart rate estimations.

Utilizing the dataset they collected, \cite{7} demonstrates the usage of the EVM approach with the forehead region as ROI and compares the results of the algorithm using the entire face region as ROI. \cite{7} indicates a decline in heart rate estimate performance when the entire face is used and also proved that the EVM algorithm is a preprocessing that provides quality to heart rate estimation since its absence generated a larger RMSE.

The goal of \cite{8}'s study was to detect participants' emotions using estimated heart rate and other features extracted from a facial video. The authors reported that the algorithm developed to estimate heart rate had an error of six BPM during testing using their dataset. Also, they did not give any comparison of the findings of their algorithm to estimate heart rate.

\begin{figure*}[ht]
	\centering
		\includegraphics[scale=.65]{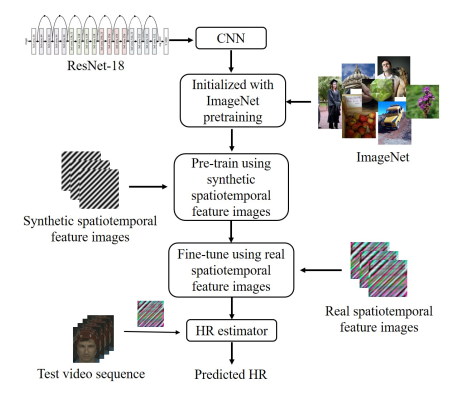}
	\caption{\cite{11} pipeline.}
	\label{FIG:11_feature_image}
\end{figure*}

In \cite{11}, the proposed method's results were evaluated twice: once using triple cross-validation on the test samples of the dataset MANHOB-HCI and once using cross-validation between datasets, utilizing the datasets VIPL-HR and UBFC-RPPG for training and actual samples of MANHOB-HCI for testing. Its pipeline is illustrated in Figure \ref{FIG:11_feature_image}. Among the metrics, the proposal outperformed the comparable methods, with the exception of MAE, which it lost to Deep-Phys.

Two experiments were carried out in \cite{12}. In the first test, just the VIPL-HR dataset was utilized for training and testing. In comparison to the other models, the suggested model outperformed them all, with the exception of M\textsubscript{e} and Pearson, where it came in second. VIPL-HR was utilized for training and MMSE-HR was used for testing in the second test where it was best.

During tests using a dataset collected by them, alternative networks were proposed in \cite{14}, which were employed in various conditions such as no movement of the subject's face, movement of the subject's face, or capturing video from cameras.
When the face was not moving, the Combined Loss (CL) model achieved a higher MAE and Cover accuracy. When the subject's face moved, the CL+F model with Filtering layers achieved higher MAE and Cover values than the other models. Furthermore, in testing with Cam\textsubscript{1} and Cam\textsubscript{2}, the CL+F model scored the best in terms of metrics. For Cam\textsubscript{3}, the CL model achieved the best MAE and Cover results. The CL+F model produced the best results for MAE and Cover in the tests conducted with the whole dataset. The estimations of the proposed model were compared to those of a medical device for heartbeat extraction in \cite{17}, and the authors found that the accuracy was high.

In \cite{18}, the proposed model outperformed the ICA and Independent Vector Analysis (IVA) approaches in heart rate estimation, outperforming the ICA and IVA methods employed in the comparison.

Table \ref{Table:tabelametricas} shows the metrics mentioned by the reviewed studies. 

\begin{table*}[ht]
\centering
\begin{tabular}{cc}
\hline
Metric                                    & Ref.                                                                        \\ \hline
Mean Error           & \cite{1}\cite{2}\cite{4}\cite{12}                                           \\ \hline
Standard Deviation  & \cite{1}\cite{2}\cite{3}\cite{4}\cite{11}\cite{12}\cite{18}                 \\ \hline
Root Mean Squared Error            & \cite{1}\cite{2}\cite{3}\cite{4}\cite{6}\cite{7}\cite{11}\cite{12}\cite{18} \\ \hline
Mean Error Rate                     & \cite{1}\cite{2}\cite{4}\cite{11}\cite{12}\cite{18}                         \\ \hline
Pearson Correlation                       & \cite{1}\cite{2}\cite{4}\cite{6}\cite{11}\cite{12}                          \\ \hline
Mean Absolute Error                 & \cite{3}\cite{5}\cite{6}\cite{11}\cite{12}\cite{14}                         \\ \hline
Mean Absolute Percentage Error (MAPE)     & \cite{3}                                                                    \\ \hline
Pearson Product Moment Correlation & \cite{3}                                                                    \\ \hline
Signal-to-Noise Ratio               & \cite{5}\cite{6}                                                            \\ \hline
Coverage at ±3 bpm                        & \cite{14}                                                                   \\ \hline
\end{tabular}
\caption{Table with the metrics used by the articles.}
\label{Table:tabelametricas}
\end{table*}


We found that Mean Error, Standard Deviation, Root Mean Squared Error, Mean Error Rate, Pearson Correlation, and Mean Absolute Error are the most often metrics mentioned by the reviewed articles, while many contain combinations of metrics assessed. We also observed that datasets are used differently and various approaches, such as super-resolution, eulerian video magnification, and Combined Loss, are used to enhance heart rate estimation.

\section{Conclusions} \label{Conclusions}

Our review examined techniques and models that extract the recorded individual's heart rate, the face as the best ROIs for extraction of heart rate signals, and the methods of processing, as well as the datasets relevant to the training of these models. We observed that the majority of datasets are for academic purposes and require approval from the authors. Furthermore, the vast majority of them provide RGB video of the participant's face, as well as heart rate, PPG, or ECG data. We also discovered that using DL models in conjunction with approaches for removing noise from movies improves performance for heart rate estimation. Regression metrics such as Mean Error, Standard Deviation, RMSE, MAE, MER, and Pearson Correlation were identified to validate the models' performance.

In terms of future research, we suggest testing the models with the identified datasets to investigate whether the models produce better outcomes in estimating heart rate in face videos.

\bibliographystyle{unsrt}  
\bibliography{references}

\end{document}